\documentclass[10pt,twocolumn,letterpaper]{article}
 \pdfoutput=1
\usepackage{iccv}
\usepackage{times}
\usepackage{epsfig}
\usepackage{graphicx}
\usepackage{amsmath}
\usepackage{amssymb}
\newcommand{\improv}[1]{\color[rgb]{0,0.8,0}\textbf{#1}}

\usepackage{gensymb}
\usepackage{makecell}
\usepackage{caption}
\usepackage{multirow}
\usepackage{multicol}
\usepackage{xcolor}
\usepackage{tabularx}
\usepackage{placeins}
\usepackage{adjustbox}
\usepackage{tabularx}
\usepackage[accsupp]{axessibility} 

\usepackage[breaklinks=true,bookmarks=false]{hyperref}
\usepackage{cleveref}
\definecolor{mypink1}{rgb}{0.858, 0.188, 0.478}
\iccvfinalcopy 

\def\iccvPaperID{20} 

\ificcvfinal\fi

\let\oldtwocolumn\twocolumn
\renewcommand\twocolumn[1][]{%
    \oldtwocolumn[{#1}{
    \begin{center}
           \includegraphics[width=\textwidth]{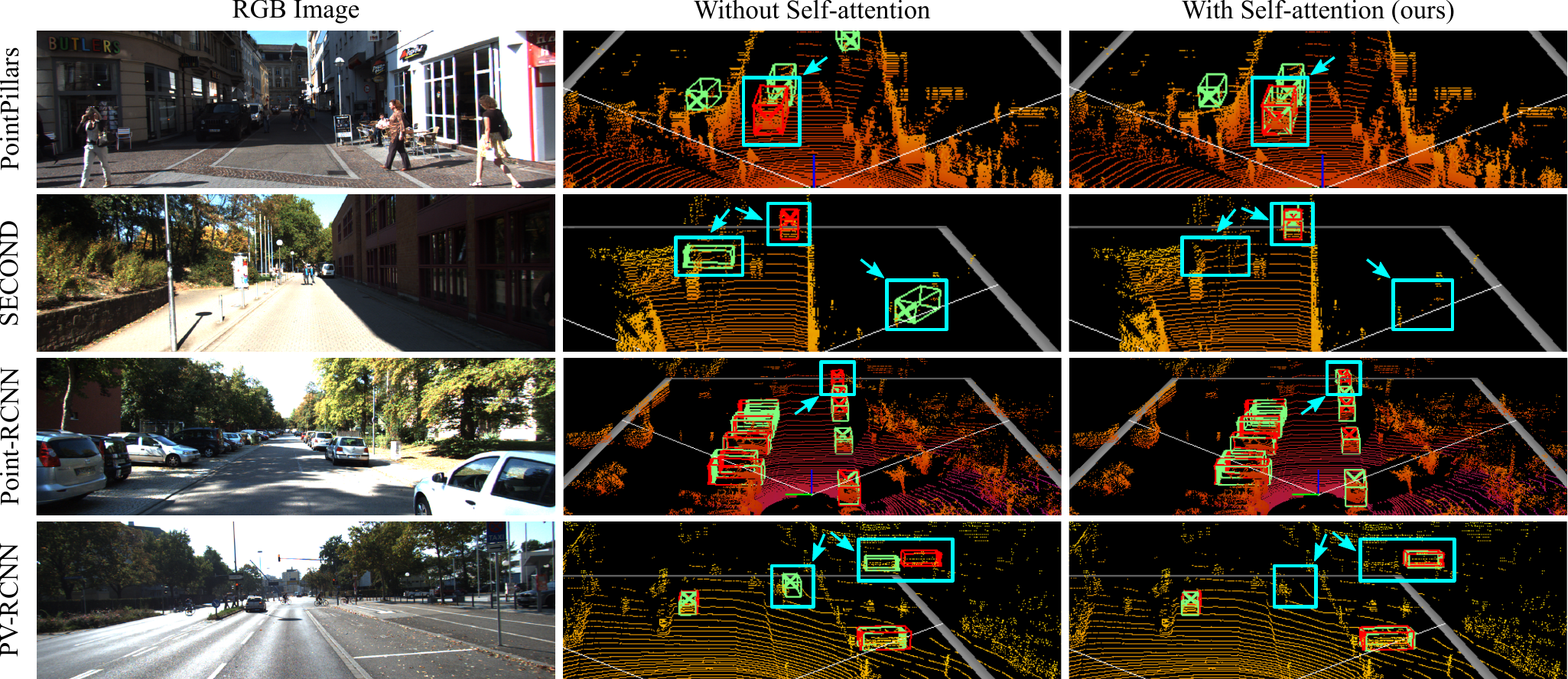}
        \end{center}
        \captionof{figure}{Performance illustrations on KITTI \textit{val}. Red bounding box is ground truth, green is detector outputs. From left to right: (a) RGB images (b)  Result of state-of-the-art methods: PointPillars \cite{pointpillars}, SECOND \cite{SECOND}, Point-RCNN \cite{PointRCNN} and PV-RCNN \cite{PVRCNN}. (c) Result of our full self-attention (FSA) augmented baselines. Our method identifies missed detections and removes false positives.
    }
    \label{fig:teaser}
    \vspace{0.3cm}
    }]
}

\begin{document}

\title{SA-Det3D: Self-Attention Based Context-Aware 3D Object Detection}

\author{
Prarthana Bhattacharyya \hspace{0.5cm}  Chengjie Huang \hspace{0.5cm}  Krzysztof Czarnecki\\
University of Waterloo, Canada\\
{\tt\small \{p6bhatta, c.huang, k2czarne\}@uwaterloo.ca}
}
\maketitle

\begin{abstract}
\vspace{-0.3cm}
Existing point-cloud based 3D object detectors use convolution-like operators to process information in a local neighbourhood with fixed-weight kernels and aggregate global context hierarchically. However, non-local neural networks and self-attention for 2D vision have shown that explicitly modeling long-range interactions can lead to more robust and competitive models. In this paper, we propose two variants of self-attention for contextual modeling in 3D object detection by augmenting convolutional features with self-attention features. We first incorporate the pairwise self-attention mechanism into the current state-of-the-art BEV, voxel and point-based detectors and show consistent improvement over strong baseline models of up to 1.5 3D AP while simultaneously reducing their parameter footprint and computational cost by 15-80\% and 30-50\%, respectively, on the KITTI validation set. We next propose a self-attention variant that samples a subset of the most representative features by learning deformations over randomly sampled locations. This not only allows us to scale explicit global contextual modeling to larger point-clouds, but also leads to more discriminative and informative feature descriptors.  Our method can be flexibly applied to most state-of-the-art detectors with increased accuracy and parameter and compute efficiency. We show our proposed method improves 3D object detection performance on KITTI, nuScenes and Waymo Open datasets. Code is available at \textcolor{mypink1}{\url{https://github.com/AutoVision-cloud/SA-Det3D}}.
\vspace{-0.5cm}
\end{abstract}

\setlength{\tabcolsep}{1.7pt}
\begin{table*}[t]
\footnotesize
    \centering
    \begin{tabular*}{\textwidth}{c|c|c|c|c|c|c}
        \hline
        Method & Task & Modality & Context & Scalability & \makecell{Attention + Convolution \\ Combination} & Stage Added    \\
        \hline
        HG-Net \cite{hgnet} & detection & points &  global-static & - & gating &  Attention modules are    \\
        PCAN \cite{pcan} & place-recognition & points & local-adaptive & - & gating &  added at the end. \\ 
        \hline

        Point-GNN \cite{pointgnn} & detection & points & local-adaptive & - & - & \\ 
        GAC \cite{gac} & segmentation & points & local-adaptive & - & - & Attention modules fully  \\
         PAT \cite{pat} & classification & points & global-adaptive & randomly sample points subset & - & replace convolution and \\
        ASCN \cite{ascn} & segmentation & points & global-adaptive & randomly sample points subset & - & set-abstraction layers. \\
        Pointformer \cite{pointformer} & detection & points & global-adaptive &  sample points subset and refine  & - & \\
        \hline

        MLCVNet \cite{mlcv} & detection & points & global-static & - & residual addition & \\
        TANet \cite{tanet} & detection & voxels & local-adaptive & - & gating & Attention modules are \\
        PMPNet \cite{pmpnet} & detection & pillars & local-adaptive & - & gated-recurrent-unit &  inserted into \\
        SCANet \cite{scanet} & detection & BEV & global-static & - & gating & the backbone. \\
        A-PointNet \cite{a-pointnet} & detection & points & global-adaptive & attend sequentially to small regions & gating &  \\
        \hline
        
        \makecell{\textbf{Ours} \\ (FSA/DSA)} & detection & \makecell{points, voxels, \\ pillars, hybrid} & global-adaptive & \makecell{attend to salient regions \\ using learned deformations} & residual addition & \makecell{Attention modules are \\ inserted into \\ the backbone.}\\ 
        \hline
    \end{tabular*}
    \caption{Properties of recent attention-based models for point-clouds}
    \label{tab:related_work_compare}
    \vspace{-0.3cm}
\end{table*}
\section{Introduction}
3D object detection has been receiving increasing attention in the computer vision and graphics community, driven by the ubiquity of LiDAR sensors and its widespread applications in autonomous driving and robotics. Point-cloud based 3D object detection has especially witnessed tremendous advancement in recent years \cite{pointpillars, SECOND, PointRCNN,  PVRCNN, F-PointNet, 3D-SSD, STD, sa-ssd, Voxelnet, votenet}. Grid-based methods first transform the irregular point-clouds to regular representations such as 2D bird's-eye view (BEV) maps or 3D voxels and process them using 2D/3D convolutional networks (CNNs). Point-based methods sample points from the raw point-cloud and query a local group around each sampled  point to define convolution-like operations \cite{pointnetplusplus, kpconv, pointconv} for point-cloud feature extraction.
\begin{figure*}[t]
    \centering
    \includegraphics[height=5.5cm]{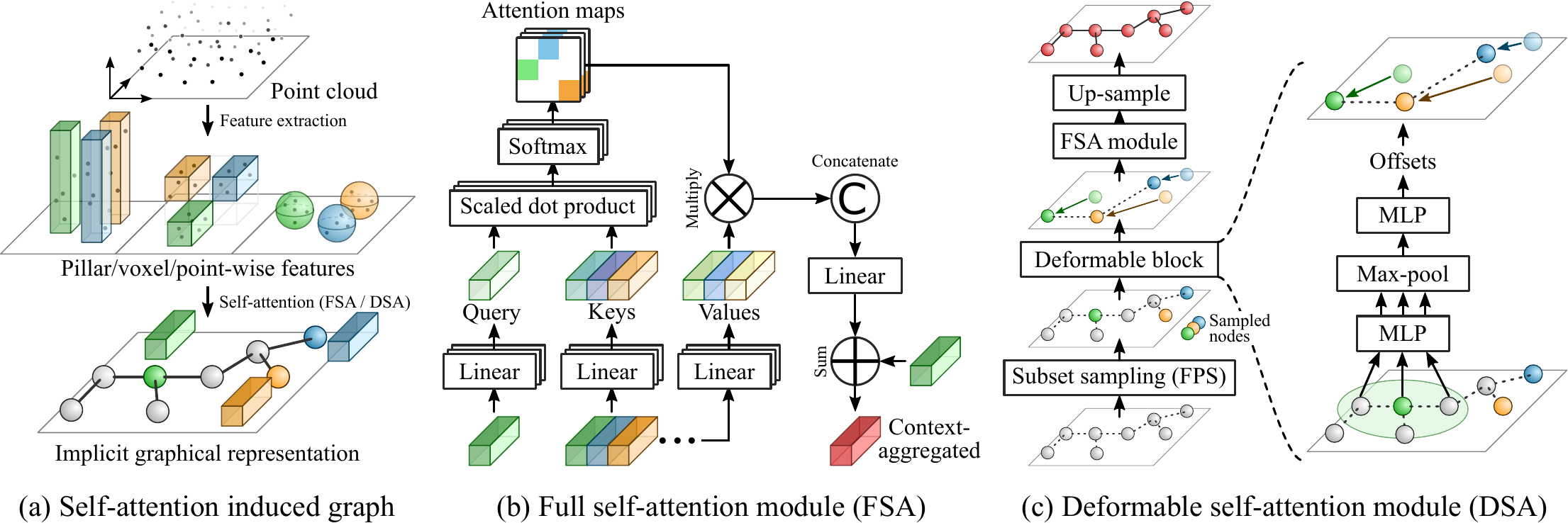}
    \caption{Architectures of the proposed FSA and DSA modules.}
    \label{fig:proposed}
\end{figure*}
Both 2D/3D CNNs and point-wise convolutions process a local neighbourhood and aggregate global context by applying feature extractors hierarchically across many layers. This has several limitations: the number of parameters scales poorly with increased size of the receptive field; learned filters are stationary across all locations; and it is challenging to coordinate the optimization of parameters across multiple layers to capture patterns in the data \cite{sagan}. 

In addition, point-cloud based 3D object detectors have to deal with missing/noisy data and a large imbalance in points for nearby and faraway objects. This motivates the need for a feature extractor that can learn global point-cloud correlations to produce more powerful, discriminative and robust features. For example, there is a strong correlation between the orientation features of cars in the same lane and this can be used to produce more accurate detections especially for distant cars with fewer points. High-confidence false positives produced by a series of points that resemble a part of an object can be also be eliminated by adaptively acquiring context information at increased resolutions.

Self-attention \cite{SA} has recently emerged as a basic building block for capturing long-range interactions. The key idea of self-attention is to acquire global information as a weighted summation of features from all positions to a target position, where the corresponding weight is calculated \textit{dynamically} via a similarity function between the features in an embedded space at these positions. The number of parameters is independent of the scale at which self-attention processes long-range interactions. Inspired by this idea, we propose two self-attention based context-aware modules to augment the standard convolutional features---Full Self-Attention (FSA) and Deformable Self-Attention (DSA). Our FSA module computes pairwise interactions among all non-empty 3D entities, and the DSA module scales the operation to large point-clouds by computing self-attention on a representative and informative subset of features. Our experiments show that we can improve the performance of current 3D object detectors with our proposed FSA/DSA blocks while simultaneously promoting parameter and compute efficiency.
\vspace{0.3cm}
\\ \textbf{Contributions}
\begin{itemize}
  \item We propose a generic globally-adaptive context aggregation module that can be applied across a range of modern architectures including BEV \cite{pointpillars}, voxel \cite{SECOND}, point \cite{PointRCNN} and point-voxel \cite{PVRCNN} based 3D detectors. We show that we can outperform strong baseline implementations by up to 1.5 3D AP (average precision) while simultaneously reducing parameter and compute cost by 15-80\% and 30-50\%, respectively, on the KITTI validation set.
  \vspace{-0.4cm}
  \item We design a scalable self-attention variant that learns to deform randomly sampled locations to cover the most representative and informative parts and aggregate context on this subset. This allows us to aggregate global context in large-scale point-clouds like nuScenes and Waymo Open dataset. 
  \vspace{-0.4cm}
  \item Extensive experiments demonstrate the benefits of our proposed FSA/DSA modules by consistently improving the performance of state-of-the-art detectors on KITTI \cite{KITTI}, nuScenes \cite{nuscenes} and Waymo Open dataset \cite{waymo}.
\end{itemize}

\section{Related Works}
\paragraph{3D Object Detection} Current 3D object detectors include BEV, voxel, point or hybrid (point-voxel) methods.
\\\textit{BEV-based} methods like MV3D \cite{MV3D} fuse multi-view representations of the point-cloud and use 2D convolutions for 3D proposal generation. PointPillars \cite{pointpillars} proposes a more efficient BEV representation and outperforms most fusion-based approaches while being 2-4 times faster. \textit{Voxel-based} approaches, on the other hand, divide the point-cloud into 3D voxels and process them using 3D CNNs \cite{Voxelnet}. SECOND \cite{SECOND} introduces sparse 3D convolutions for efficient 3D processing of voxels, and CBGS \cite{cbgs} extends it with multiple heads. \textit{Point-based} methods are inspired by the success of PointNet \cite{pointnet} and PointNet++ \cite{pointnetplusplus}. F-PointNet \cite{F-PointNet}  first applied PointNet for 3D detection, extracting point-features from point-cloud crops that correspond to 2D camera-image detections. Point-RCNN \cite{PointRCNN} segments 3D point-clouds using PointNet++, and uses the segmentation features to better refine box proposals. \textit{Point-Voxel-based} methods like STD \cite{STD}, PV-RCNN \cite{PVRCNN} and SA-SSD \cite{sa-ssd} leverage both voxel and point-based abstractions to produce more accurate bounding boxes. 
\\ \textbf{Relationship to current detectors:} Instead of repeatedly stacking convolutions, we propose a simple, scalable, generic and permutation-invariant block called FSA/DSA to adaptively aggregate context information from the entire point-cloud. This allows remote regions to directly communicate and can help in learning relationships across objects. This module is flexible and can be applied in parallel to convolutions within the backbone of modern point-cloud based detector architectures. 

\vspace{-0.5cm}
\paragraph{Attention for Context Modeling}
Self-attention \cite{SA} has been instrumental to achieving state-of-the-art results in machine translation and combining self-attention with convolutions is a theme shared by recent work in natural language processing \cite{NLP-SA}, image recognition \cite{AA}, 2D object detection \cite{AttnUNet}, activity recognition \cite{NL}, person re-identification  \cite{Reid} and reinforcement learning \cite{RL}. 
\\ Using self-attention to aggregate global structure in point-clouds for 3D object detection remains a relatively unexplored domain. PCAN \cite{pcan}, TANet \cite{tanet}, Point-GNN \cite{pointgnn}, GAC \cite{gac}, PMPNet \cite{pmpnet} use local context to learn context-aware discriminative features. However relevant contextual information can occur anywhere in the point-cloud and hence we need global context modeling. HGNet \cite{hgnet}, SCANet \cite{scanet}, MLCVNet \cite{mlcv} use global scene semantics to improve performance of object detection, but the global context vector is shared across all locations and channels and does not adapt itself according to the input features leading to a sub-optimal representation.
PAT \cite{pat}, ASCN \cite{ascn}, Pointformer \cite{pointformer} build globally-adaptive point representations for classification, segmentation and 3D detection. But because they use the costly pairwise self-attention mechanism, the self-attention does not scale to the entire point-cloud. Consequently, they process a randomly selected subset of points, which may be sensitive to outliers. To process global context for 3D object detection and scale to large point-clouds, Attentional PointNet \cite{a-pointnet}  uses GRUs \cite{GRU} to sequentially attend to different parts of the point-cloud. Learning global context by optimizing the hidden state of a GRU is slow and inefficient, however. 
\\ In contrast, our method can processes context \emph{adaptively} for each location from the entire point-cloud, while also \emph{scaling} to large sets using learned deformations. Since the global context is fused with local-convolutional features, the training is stable and efficient as compared to GRUs or stand-alone attention networks \cite{Stand-alone}. \Cref{tab:related_work_compare} compares our work with recent point-cloud based attention methods.
\begin{figure*}[t]
    \centering
    \includegraphics[width=0.825\textwidth]{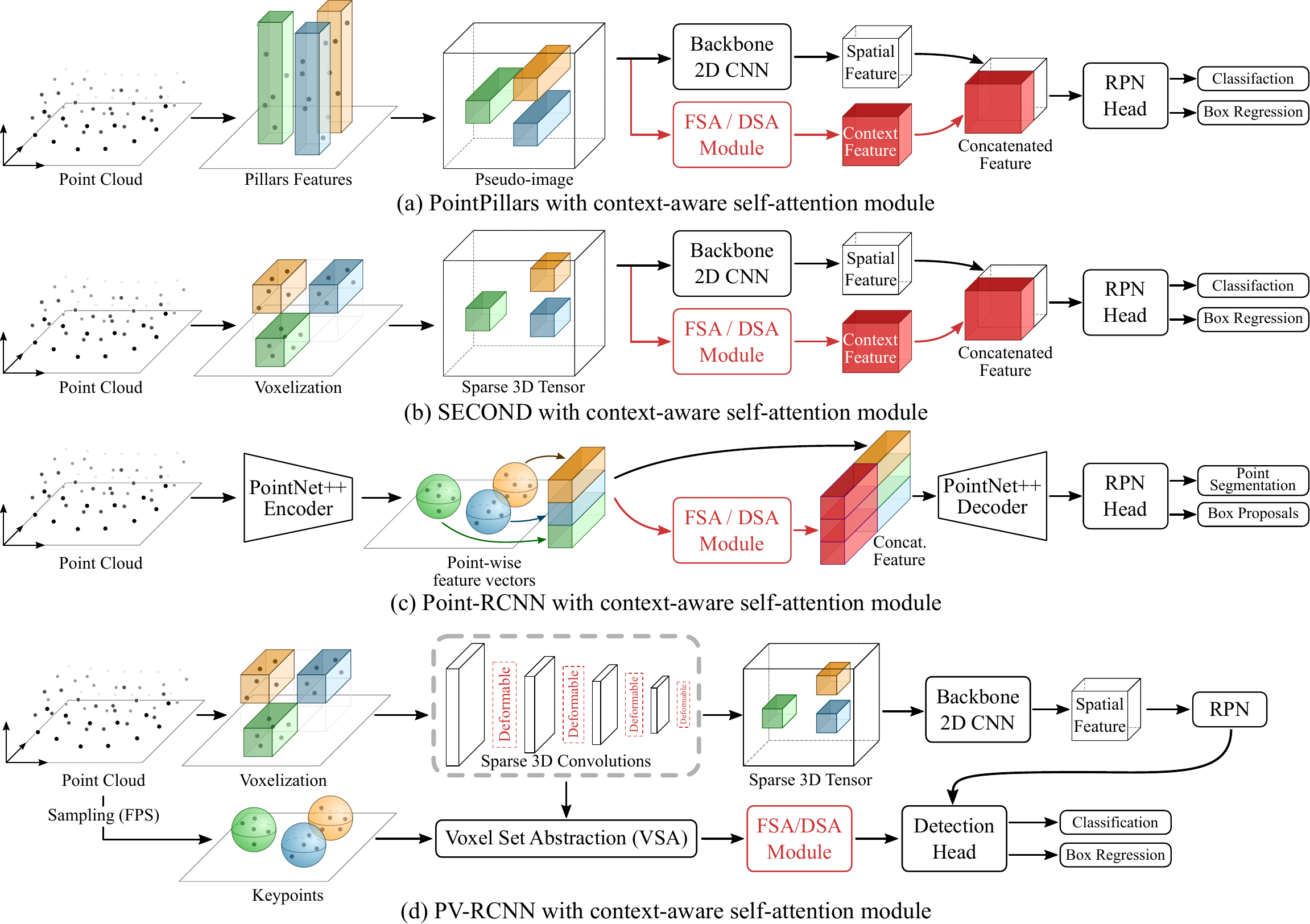}
    \caption{Proposed FSA/DSA module augmented network architectures for different backbone networks.}
    \label{fig:architectures}
\end{figure*}
\section{Methods}
In this section, we first introduce a Full Self-Attention (FSA) module for discriminative feature extraction in 3D object detection that aims to produce more powerful and robust representations by exploiting global context. Next, inspired by 2D deformable convolutions \cite{DCN} we introduce a variant of FSA called Deformable Self-Attention (DSA). DSA can reduce the quadratic computation time of FSA and scale to larger and denser point-clouds.
The two proposed modules are illustrated in \Cref{fig:proposed}.

\setlength{\tabcolsep}{1pt}
\begin{table*}[t]
\footnotesize
    \centering
    \begin{tabular*}{\textwidth}{ @{\extracolsep{\fill}}c|cccc|cccc|cccc|cccc}
        \hline
        \multirow{2}{*}{Method} & 
        \multicolumn{4}{c|}{PointPillars \cite{pointpillars}} &
        \multicolumn{4}{c|}{SECOND \cite{SECOND}} &
        \multicolumn{4}{c|}{Point-RCNN \cite{PointRCNN}} &
        \multicolumn{4}{c}{PV-RCNN \cite{PVRCNN}} \\
        & 3D & BEV & Param & FLOPs &
        3D & BEV & Param & FLOPs &
        3D & BEV & Param & FLOPs &
        3D & BEV & Param & FLOPs\\
        \hline
        Baseline &
        78.39 & 88.06 & 4.8 M & 63.4 G & 
        81.61 & 88.55 & 4.6 M & 76.9 G & 
        80.52 & \textbf{88.80} & 4.0 M & 27.4 G &
        84.83 & \textbf{91.11} & 12 M & 89 G \\
        
        DSA &
        78.94 & 88.39 & 1.1 M & 32.4 G &
        \textbf{82.03} & 89.82 & 2.2 M & 52.6 G &
        81.80 & 88.14 & 2.3 M & 19.3 G &
        84.71 & 90.72 & 10 M & 64 G \\
        
        FSA &
        \textbf{79.04} & \textbf{88.47} & 1.0 M & 31.7 G &
        81.86 & \textbf{90.01} & 2.2 M & 51.9 G &
        \textbf{82.10} & 88.37 & 2.5 M & 19.8 G &
        \textbf{84.95} & 90.92 & 10 M & 64.3 G \\
        \hline
        
        \textit{Improve.} &
        \improv{+0.65} & \improv{+0.41} & \improv{-79\%} & \improv{-50\%} &
        \improv{+0.42} & \improv{+1.46} & \improv{-52\%} & \improv{-32\%} &
        \improv{+1.58} & - & \improv{-37\%} & \improv{-38\%} &
        \improv{+0.12} & - & \improv{-16\%} & \improv{-27\%}
        \\
        \hline
    \end{tabular*}
    \caption{Performance comparison for moderate difficulty Car class on KITTI \textit{val} split with 40 recall positions}
    \label{tab:kitti_val_car}
    \vspace{-0.3cm}
\end{table*}

\subsection{Formulation}
For the input set $\mathcal{X}=\{\textbf{x}_1, \textbf{x}_2,...\textbf{x}_n\}$ of $n$ correlated features and $i \in \{1,...n\}$, we propose to use self-attention introduced by Vaswani et al. \cite{SA} to exploit the pairwise similarities of the $i^{th}$ feature node with all the feature nodes, and stack them to compactly represent the global structural information for the current feature node. 

Mathematically, the set of pillar/voxel/point features and their relations are denoted by a graph $G= (\mathcal{V},\mathcal{E})$, which comprises the node set $\mathcal{V} = \{\textbf{x}_1, \textbf{x}_2,...\textbf{x}_n \in R^d \}$, together with an edge set $\mathcal{E} = \{\textbf{r}_{i,j} \in R^{N_h}, i= 1,...,n$ and $j= 1,...,n\}$. A self-attention module takes the set of feature nodes, and computes the edges (see \Cref{fig:proposed}\,(a)). The edge $\textbf{r}_{i,j}$ represents the relation between the $i^{th}$ node and the $j^{th}$ node, and $N_h$ represents the number of heads (number of attention maps in \Cref{fig:proposed}\,(b)) in the attention mechanism across $d$ feature input channels as described below. We assume that $N_h$ divides $d$ evenly. The advantage of representing the processed point-cloud features as nodes in a graph is that now the task of aggregating global context is analogous to capturing higher order interaction among nodes by message passing on graphs for which many mechanisms like self-attention exist. 

\subsection{Full Self-Attention Module}
Our Full Self-Attention (FSA) module projects the features $\textbf{x}_i$ through linear layers into matrices of query vectors Q, key vectors K, and value vectors V (see \Cref{fig:proposed}(b)). The similarities between query $\textbf{q}_i$ and all keys, $\textbf{k}_{j=1:n}$, are computed by a dot-product, and normalized into attention weights $\textbf{w}_i$, via a softmax function. The attention weights are then used to compute the pairwise interaction terms, $\textbf{r}_{ij} = w_{ij}\textbf{v}_j$. The accumulated global context for each node vector $\textbf{a}_i$ is the sum of these pairwise interactions, $\textbf{a}_i =\sum_{j=1:n} \textbf{r}_{ij}$. As we mentioned in our formulation, we also use multiple attention heads, applied in parallel, which can pick up channel dependencies independently. The final output for the node $i$ is then produced by concatenating the accumulated context vectors $\textbf{a}_i^{h=1:N_h}$ across heads, passing it through a linear layer, normalizing it with group normalization \cite{group} and summing it with $\textbf{x}_i$ (residual connection).

\textbf{Advantages:} The important advantage of this module is that the resolution at which it gathers context is independent of the number of parameters and the operation is permutation-invariant. This makes it attractive to replace a fraction of the parameter-heavy convolutional filters at the last stages of 3D detectors with self-attention features for improved feature quality and parameter efficiency.

\textbf{Complexity:} The pairwise similarity calculation is $\mathcal{O}(n^2d)$ in nature. The inherent sparsity of point-clouds and the efficient matrix-multiplication based pairwise computation makes FSA a viable feature extractor in current 3D detection architectures. However, it is necessary to trade accuracy for computational efficiency in order to scale to larger point-clouds. In the next section, we propose our Deformable Self-Attention module to reduce the quadratic computation time of FSA.

\subsection{Deformable Self-Attention Module}
 Our primary idea is to attend to a representative subset of the original node vectors in order to aggregate global context. We then up-sample this accumulated structural information back to all node locations. We describe the up-sampling process in the supplementary section. The complexity of this operation is $\mathcal{O}(m^2d)$, where $m << n$ is the number of points chosen in the subset. In order for the subset to be representative, it is essential to make sure that the selected nodes cover the informative structures and common characteristics in 3D geometric space. Inspired by deformable convolution networks \cite{DCN} in vision, we propose a geometry-guided vertex refinement module that makes the nodes self-adaptive and spatially recomposes them to cover locations which are important for semantic recognition. Our node offset-prediction module is based on vertex alignment strategy proposed for domain alignment \cite{Pointdan, meshrcnn}. Initially $m$ nodes are sampled from the point-cloud by farthest point sampling (FPS) with vertex features $\textbf{x}_i$ and a 3D vertex position $v_i$. For the $i^{th}$ node, the updated position $v'_i$ is calculated by aggregating the local neighbourhood features with different significance as follows:
\begin{equation}
x^*_i = \frac{1}{k} \text{ReLU} \sum_{j \in \mathcal{N}(i)}{W_{\text{offset}}(\textbf{x}_i-\textbf{x}_j)\cdot(v_i-v_j)} 
\end{equation}
\begin{equation}
v'_i = v_i+ \tanh(W_\text{align}x^*_i)
\end{equation}
where $\mathcal{N}_i$ gives the $i$-th node's $k$-neighbors in the point-cloud and $W_\text{offset}$ and $
W_\text{align}$ are weights learned end-to-end. The final node features are computed by a non-linear processing of the locally aggregated embedding as follows:
\begin{equation}
\textbf{x}'_i = \max_{j \in \mathcal{N}(i)}W_{out}\textbf{x}_j 
\end{equation}
Next, the $m$ adaptively aggregated features $\{\textbf{x}'_1....\textbf{x}'_m\}$ are then passed into a full self-attention (FSA) module to model relationships between them.
This aggregated global information is then shared among all $n$ nodes from the $m$ representatives via up-sampling. We call this module a Deformable Self-Attention (DSA) module as illustrated in \Cref{fig:proposed}(c). 

\textbf{Advantages:} The main advantage of DSA is that it can scalably aggregate global context for pillar/voxel/points. Another advantage of DSA is that it is trained to collect information from the most informative regions of the point-cloud, improving the feature descriptors. 

\section{Experiments}
\setlength{\tabcolsep}{7.6pt}
\begin{table*}[t]
    \centering
    \footnotesize
    \begin{tabular*}{\textwidth}{c||ccc|ccc|ccc|ccc}
        \hline
        \multirow{2}{*}{Model} &
        \multicolumn{3}{c|}{Car - 3D} &
        \multicolumn{3}{c|}{Car - BEV} &
        \multicolumn{3}{c|}{Cyclist - 3D} &
        \multicolumn{3}{c}{Cyclist - BEV}  \\
        & Easy & Mod. & Hard &
        Easy & Mod. & Hard &
        Easy & Mod. & Hard &
        Easy & Mod. & Hard \\
        \hline

        MV3D \cite{MV3D} & 
        74.97 & 63.63 & 54.00 &
        86.62 & 78.93 & 69.80 &
        - & - & - & - & - & - \\


        
        
        
        PointPillars \cite{pointpillars} & 
        82.58 & 74.31 & 68.99 & 
        90.07 & 86.56 & 82.81 & 
        77.10 & 58.65 & 51.92 &
        79.90 & 62.73 & 55.58 \\
        
        SECOND \cite{SECOND} & 
        83.34 & 72.55 & 65.82 &
        89.39 & 83.77 & 78.59 &
        71.33 & 52.08 & 45.83 &
        76.50 & 56.05 & 49.45 \\
        
        PointRCNN \cite{PointRCNN} & 
        86.96 & 75.64 & 70.70 &
        92.13 & 87.39 & 82.72 &
        74.96 & 58.82 & 52.53 &
        82.56 & 67.24 & 60.28 \\
        
        STD \cite{STD} & 
        87.95 & 79.71 & 75.09 &
        94.74 & 89.19 & \textbf{86.42} &
        78.69 & 61.59 & 55.30 &
        81.36 & 67.23 & 59.35 \\
        
        
        
        3DSSD \cite{3D-SSD} & 
        88.36 & 79.57 & 74.55 &
        92.66 & 89.02 & 85.86 &
        \textbf{82.48} & 64.10 & 56.90 & 
        \textbf{85.04} & 67.62 & 61.14 \\
        
        SA-SSD \cite{sa-ssd} & 
        88.75 & 79.79 & 74.16 &
        \textbf{95.03} & \bf{91.03} & 85.96 &
        - & - & - & - & - & - \\
        
        TANet \cite{tanet} & 
        83.81 & 75.38 & 67.66 &
        - & - & - &
        73.84 & 59.86 & 53.46 &
        - & - & - \\
        
        Point-GNN \cite{pointgnn} &
        88.33 & 79.47 & 72.29 &
        93.11 & 89.17 & 83.90 &
        78.60 & 63.48 & 57.08 &
        81.17 & 67.28 & 59.67 \\
        
        PV-RCNN \cite{PVRCNN} & 
        \textbf{90.25} & 81.43 & 76.82 &
        94.98 & 90.65 & 86.14 &
        78.60 & 63.71 & 57.65 &
        82.49 & 68.89 & 62.41 \\

        \hline
        PV-RCNN + DSA (Ours) & 
        88.25 & \textbf{81.46} & \textbf{76.96} &
        92.42 & 90.13 & 85.93 &
        \textcolor{blue}{82.19} & \textbf{68.54} & \textbf{61.33} &
        \textcolor{blue}{83.93} & \textbf{72.61} & \textbf{65.82} \\
        
        
        \hline
    \end{tabular*}
    \caption{Performance comparison of 3D detection on KITTI \textit{test} split with AP calculated with 40 recall positions. The \textbf{best} and \textcolor{blue}{second-best} performances are highlighted across all datasets.}
    \label{tab:kitti_test}
\end{table*}
\setlength{\tabcolsep}{5.7pt}
\begin{table*}[t]
\centering
\footnotesize
    \begin{tabular*}{\textwidth}{c|c|cc|cccccccccc}
        \hline
        Model & Mode & mAP & NDS & Car & Truck & Bus & Trailer & CV &
        Ped & Moto & Bike & Tr. Cone & Barrier \\
        \hline
        
        PointPillars \cite{pointpillars} & 
        Lidar & 30.5 & 45.3 & 
        68.4 & 23.0 & 28.2 & 23.4 & 4.1 &
        59.7 & 27.4 & 1.1 & 30.8 & 38.9 \\
        
        WYSIWYG \cite{wysiwyg} & 
        Lidar & 35.0 & 41.9 & 79.1 & 30.4 & 46.6 & 40.1 & 
        7.1 & 65.0 & 18.2 & 0.1 & 28.8 & 34.7 \\
        
        PointPillars+ \cite{pointpainting} & 
        Lidar & 40.1 & 55.0&
        76.0 & 31.0 & 32.1 & 36.6 & 11.3 &
        64.0 & 34.2 & 14.0  & 45.6 & 56.4 \\
        
        PMPNet \cite{pmpnet} & 
        Lidar & 45.4 & 53.1 & 79.7 & 33.6 & 47.1 & 43.0 & 
        \textbf{18.1} & \textbf{76.5} & 40.7 & 7.9 & 58.8 & 48.8 \\
        
        SSN \cite{ssn} & 
        Lidar & 46.3 & 56.9 & 
        80.7 & 37.5 & 39.9 & 43.9 & 14.6 & 72.3 & \textbf{43.7} & 20.1 & 54.2 & 56.3 \\
        
        Point-Painting \cite{pointpainting} & 
        RGB + Lidar & 46.4 & 58.1 & 
        77.9 & 35.8 & 36.2 & 37.3 & 15.8 & 73.3 & 41.5 & \textbf{24.1} & \textbf{62.4} & \textbf{60.2} \\
        
        \hline
        PointPillars + DSA (Ours) &
        Lidar & \textbf{47.0} & \textbf{59.2} &
        \textbf{81.2} & \textbf{43.8} & \textbf{57.2} & \textbf{47.8} & 11.3 & 
        \textcolor{blue}{73.3} & 32.1 & 7.9 & \textcolor{blue}{60.6} & 55.3 \\
    
        
        \hline
    \end{tabular*}
    \caption{Performance comparison of 3D detection with PointPillars backbone on nuScenes \textit{test} split. “CV”, ”Ped” , “Moto”, “Bike”, “Tr. Cone” indicate construction vehicle, pedestrian, motorcycle, bicycle and traffic cone respectively. 
    The values are taken from the official evaluation server \small \textcolor{mypink1}{\url{https://eval.ai/web/challenges/challenge-page/356/leaderboard/1012}}.}
    \label{tab:nuscenes_test_class}
    \vspace{-0.5cm}
\end{table*}
\setlength{\tabcolsep}{8pt}
\begin{table}[t]
\footnotesize
\begin{center}
\begin{tabular}{c||c||c|c}
\hline 
Difficulty  & Method & \multicolumn{2}{c}{Vehicle} \\
\cline{3-4}
 &   &  3D AP & 3D APH \\
\hline
& StarNet \cite{starnet} & 53.7 & - \\
& PointPillars \cite{pointpillars} & 56.6 & -  \\
& PPBA \cite{ppba} & 62.4 & -  \\
& MVF \cite{MV3D} & 62.9 & -  \\
L1 & AFDet \cite{afdet} & 63.7 & -  \\
& CVCNet \cite{cvcnet} & 65.2 & -  \\
& Pillar-OD \cite{pillar-od} & 69.8 & -  \\
& $\dagger$SECOND \cite{SECOND} & 70.2 & 69.7 \\
& PV-RCNN \cite{PVRCNN} & 70.3 & 69.7 \\ 
& SECOND + DSA (Ours) & \textbf{71.1} & \textbf{70.7} \\

\hline
L2 & $\dagger$SECOND \cite{SECOND} & 62.5 & 62.0 \\
& PV-RCNN \cite{PVRCNN} & \textbf{65.4} & \textbf{64.8} \\
& SECOND + DSA (Ours) & \textcolor{blue}{63.4} & \textcolor{blue}{63.0} \\
\hline
\end{tabular}
\end{center}
\vspace{-0.2cm}
\caption{Comparison on Waymo Open Dataset \textit{validation} split for 3D vehicle detection. Our DSA model has $\textcolor{red}{52\%}$ fewer parameters and $\textcolor{red}{32\%}$ fewer FLOPs compared to SECOND and $\textcolor{red}{80\%}$ fewer parameters and $\textcolor{red}{41\%}$ fewer FLOPs compared to PV-RCNN. $\dagger$Re-implemented by \cite{OpenPCDet}}
\label{tab:waymo_val}
\vspace{-0.5 cm}
\end{table}

\subsection{Network Architectures} 
We train and evaluate our proposed FSA and DSA modules on four state-of-the-art architecture backbones: PointPillars \cite{pointpillars}, SECOND \cite{SECOND}, Point-RCNN \cite{PointRCNN}, and PV-RCNN \cite{PVRCNN}. The architectures of the backbones are illustrated in \Cref{fig:architectures}. The augmented backbones can be trained end-to-end without additional supervision. 
\\ For the KITTI dataset, the detection range is within [0,70.4]\,m, [-40,40]\,m and [-3,1]\,m for the XYZ axes, and we set the XY pillar resolution to (0.16, 0.16)\,m and XYZ voxel-resolution of (0.05, 0.05, 0.1)\,m. For nuScenes, the range is [-50,50]\,m, [-50,50]\,m, [-5,3]\,m along the XYZ axes and the XY pillar resolution is (0.2, 0.2)\,m. For the Waymo Open dataset, the detection range is [-75.2, 75.2]\,m for the X and Y axes and [-2, 4]\,m for the Z-axis, and we set the voxel size to (0.1, 0.1, 0.15)\,m. Additionally, the deformation radius is set to 3\,m, and the feature interpolation radius is set to 1.6\,m with 16 samples. The self-attention feature dimension is 64 across all models. We apply 2 FSA/DSA modules with 4 attention heads across our chosen baselines. For DSA, we use a subset of 2,048 sampled points for KITTI and 4,096 sampled points for nuScenes and Waymo Open Dataset. We use standard data-augmentation for point clouds. For baseline models, we reuse the pre-trained checkpoints provided by OpenPCDet \cite{OpenPCDet}. More architectural details are provided in the supplementary.

\subsection{Implementation Details}
\textbf{KITTI:} KITTI benchmark \cite{KITTI} is a widely used benchmark with 7,481 training samples and 7,518 testing samples. We follow the standard split \cite{MV3D} and divide the training samples into \textit{train} and \textit{val} split with 3,712 and 3,769 samples respectively. All models were trained on 4 NVIDIA Tesla V100 GPUs for 80 epochs with Adam optimizer \cite{Adam} and one cycle learning rate schedule \cite{onecycle}.  We also use the same batch size and learning rates as the baseline models. 

\textbf{nuScenes}
nuScenes \cite{nuscenes} is a more recent large-scale benchmark for 3D object detection. In total, there are 28k, 6k, 6k, annotated frames for training, validation, and testing, respectively. The annotations include 10 classes with a long-tail distribution. We train and evaluate a DSA model with PointPillars as the backbone architecture. All previous methods combine points from current frame and previous frames within 0.5\,s, gathering about 300\,k points per frame. FSA does not work in this case since the number of pillars in a point cloud is too large to fit the model in memory. In DSA, this issue is avoided by sampling a representative subset of pillars. The model was trained on 4 NVIDIA Tesla V100 GPUs for 20 epochs with a batch size of 8 using Adam optimizer \cite{Adam} and one cycle learning rate schedule \cite{onecycle}. 

\textbf{Waymo Open Dataset} Waymo Open Dataset \cite{waymo} is currently the largest dataset for 3D detection for autonomous  driving. There are 798 training sequences with 158,081 LiDAR samples, and 202 validation sequences with 39,987 LiDAR samples. The objects are annotated in the full 360\degree field of view. We train and evaluate a DSA model with SECOND as the backbone architecture. The model was trained on 4 NVIDIA Tesla V100 GPUs for 50 epochs with a batch size of 8 using Adam optimizer \cite{Adam} and one cycle learning rate schedule \cite{onecycle}. 

\section{Results}
\subsection{3D Detection on the KITTI Dataset}
On KITTI, we report the performance of our proposed model on both \textit{val} and \textit{test} split. We focus on the average precision for moderate difficulty and two classes: car and cyclist. We calculate the average precision on \textit{val} split with 40 recall positions using IoU threshold of 0.7 for car class and 0.5 for cyclist class. The performance on \textit{test} split is calculated using the official KITTI test server.
\vspace{0.1cm}
\\ \textbf{Comparison with state-of-the-art:} \Cref{tab:kitti_val_car} shows the results for car class on KITTI \textit{val} split. For all four state-of-the-art models augmented with DSA and FSA, both variants were able to achieve performance improvements over strong baselines with significantly fewer parameters and FLOPs. On KITTI \textit{test} split, we evaluate PV-RCNN+DSA and compare it with the models on KITTI benchmark. The results are shown in \Cref{tab:kitti_test}. On the car class DSA shows an improvement of 0.15 3D AP on the hard setting, while for the smaller cyclist class we achieve significantly better performance than all other methods with upto 4.5 3D AP improvement on the moderate setting. Overall, the results consistently demonstrate that adding global contextual information benefits performance and efficiency, especially for the difficult cases with smaller number of points.

\subsection{3D Detection on the nuScenes Dataset}
To test the performance of our methods in more challenging scenarios, we evaluate PointPillars with DSA modules on the nuScenes benchmark using the official test server. In addition to average precision (AP) for each class, nuScenes benchmark introduces a new metric called nuScenes Detection Score (NDS). It is defined as a weighted sum between mean average precision (mAP), mean average errors of location (mATE), size (mASE), orientation (mAOE), attribute (mAAE) and velocity (mAVE).
\vspace{0.1cm}
\\ \textbf{Comparison with state-of-the-art:}
We first compare our PointPillars+DSA model with PointPillars+ \cite{pointpainting}, a class-balanced re-sampled version of PointPillars inspired by \cite{cbgs}. DSA achieves about 7\% improvement in mAP and 4.2\% improvement in NDS compared to PointPillars+, even for some small objects, such as pedestrian and traffic cone. Compared with other attention and fusion-based methods like PMPNet and Point-Painting, DSA performs better in the main categories of traffic scenarios such as Car, Truck, Bus and Trailer etc. Overall, our model has the highest mAP and NDS score compared to state-of-the-art PointPillars-based 3D detectors.

\subsection{3D Detection on the Waymo Open Dataset}
We also report performance on the large Waymo Open Dataset with our SECOND+DSA model to further validate its effectiveness. The objects in the dataset are split into two levels based on the number of points in a single object, where LEVEL1 objects have at-least 5 points and the LEVEL2 objects have at-least 1 point inside. For evaluation, the average precision (AP) and average precision weighted by heading (APH) metrics are used. The IoU threshold is 0.7 for vehicles. 
\vspace{0.1cm}
\\ \textbf{Comparison with the state-of-the-art:} \Cref{tab:waymo_val} shows that our method outperforms previous state-of-the-art PV-RCNN with a 0.8\%AP and 1\%APH gain for 3D object detection while having 80\% fewer parameters and 41\% fewer FLOPs on LEVEL1. This supports that our proposed DSA is able to effectively capture global contextual information for improving 3D detection  performance. Better performance in terms of APH also indicates that context helps to predict more accurate heading direction for the vehicles. On LEVEL2, we outperform the SECOND baseline by 0.9\% AP and 1.0\% APH. Overall SECOND+DSA provides the better balance between performance and efficiency as compared to PV-RCNN. The experimental results validate the generalization ability of FSA/DSA on various datasets.

\subsection{Ablation studies and analysis}
\begin{figure}[t]
    \centering
    \includegraphics[width=0.8\linewidth]{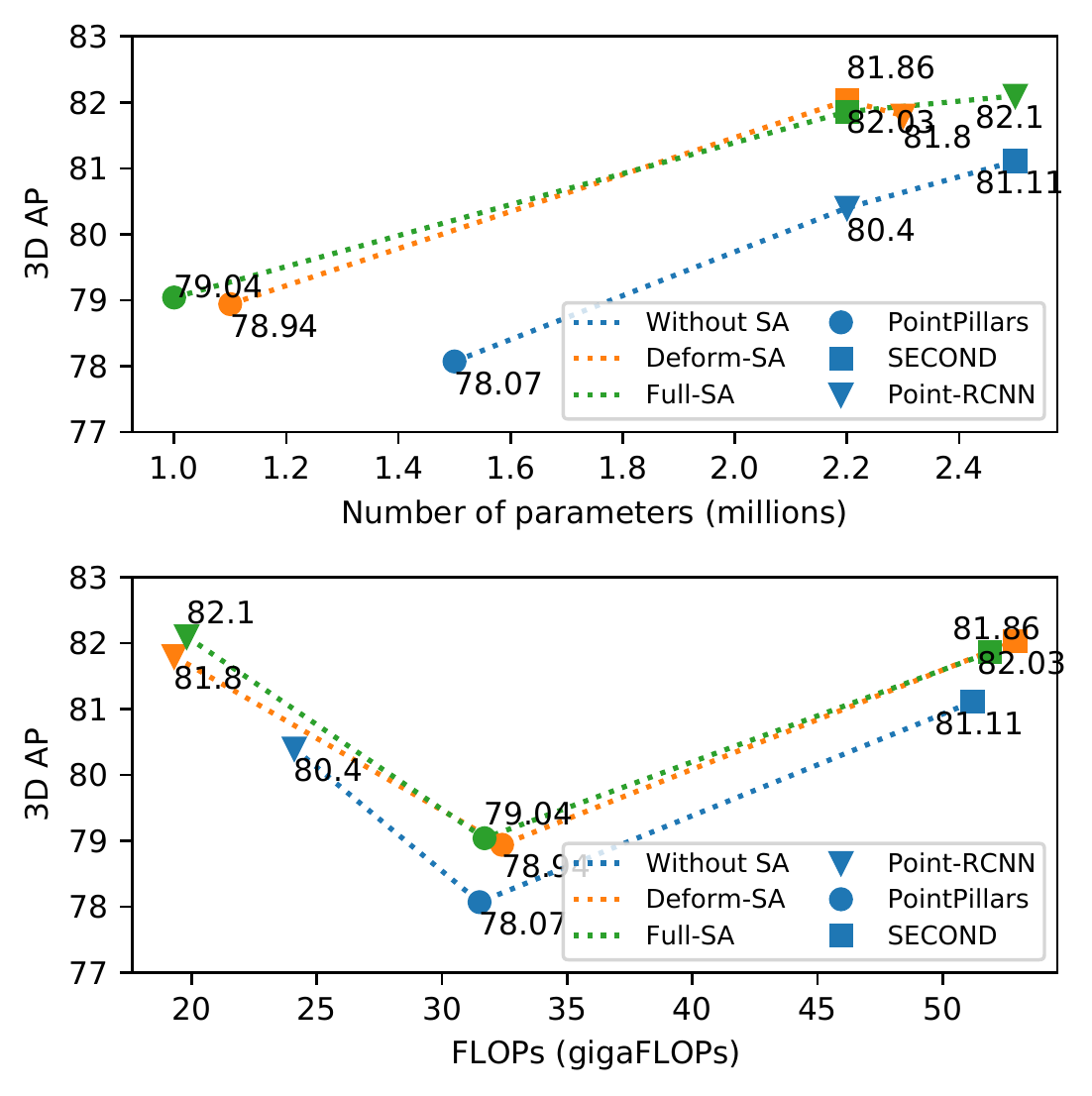}
    \caption{3D AP on moderate Car class of KITTI val split (R40) vs. number of parameters (Top) and GFLOPs (Bottom) for baseline models and proposed baseline extensions with Deformable and Full SA.}
    \label{teaser_graph}
    \vspace{-0.5cm}
\end{figure}

Ablation studies are conducted on the KITTI validation split \cite{MV3D} for moderate Car class using AP@R40, in order to validate our design choices.
\vspace{-0.4cm}
\paragraph{Model variations} In our ablation study with PointPillars backbone in \Cref{tab:ablation_components}, we represent the number of 2D convolution filters as $N_{filters}$, self-attention heads as $N_h$, self-attention layers as $N_l$, sampled points for DSA as $N_{keypts}$, deformation radius as $r_{def}$ and the up-sampling radius as $r_{up}$. 
\\ \textbf{Effect of number of filters}: We note that both FSA and DSA outperform not only the models with similar parameters by 0.97\% and 0.87\% respectively, but also the state-of-the-art models with 80\% more parameters by 0.65\% and 0.55\%. This indicates that our modules are extremely parameter efficient. Finally, we also note that if the number of parameters and compute are kept roughly the same as the baseline(Row-D), DSA outperforms the baseline by a large margin of 1.41\%. We also illustrate consistent gains in parameter and computation budget across backbones in \Cref{teaser_graph}. 
\\ \textbf{Effect of number of self-attention heads and layers} (Row-A):  We note that increasing heads from 2 to 4 leads to an improvement of 0.37\% for PointPillars. Since increasing number of self-attention layers beyond a certain value can lead to over-smoothing \cite{oversmoothing}, we use 2 FSA/DSA layers in the backbone and 4 heads for multi-head attention. 
\\ \textbf{Effect of number of sampled points} (Row-B): For DSA, we also vary the number of keypoints sampled for computation of global context. We note that the performance is relatively robust to the number of sampled points. 
\\ \textbf{Effect of deformation and upsampling radius} (Row-C): For DSA, we note that the performance is generally robust to the deformation radius upto a certain threshold, but the up-sampling radius needs to be tuned carefully. Generally an up-sampling radius of 1.6m in cars empirically works well. 
\vspace{-0.4cm}

\setlength{\tabcolsep}{1.0pt}
\begin{table}[t]
\footnotesize
    \centering
    \begin{tabular*}{\linewidth}{c||cccccc|ccc}
        \hline
        Model & $N_{filters}$ & $N_h$ & $N_l$  & $N_{keypts}$ & $r_{def}$ & $r_{up}$ & 3D AP & Params & FLOPs\\
        \hline
        baseline & (64,128,256) & - & - &  - & - & - & 78.39 & 4.8M & 63.4G \\
         & (64,64,128) & - & - &  - & - & - & 78.07 & 1.5M  & 31.5G \\
        \hline 
        & (64,64,64) & 2 & 2 &  - & - & - & 78.67 & 1.0M & 31.3G \\
        (A) &  & 4 & 1 &  - & - & - & 78.34 & 1.0M & 31.5G \\
        &  & 4 & 2 &  - & - & - & 79.04 & 1.0M & 31.7G \\
        &  & 4 & 4 &  - & - & - & 78.56 & 1.0M & 32.0G \\
        \hline
        & (64,64,64) & 4 & 2 &  512 & 3 & 1.6 & 78.70 & 1.1M & 32.4G \\
        (B) & & &  &  1024 &  &  & 78.95 & 1.1M & 32.4G \\
         &  &  &  &   2048 &  &  & 78.94 & 1.1M & 32.4G \\
        &  & &  &   4096 &  &  & 78.90 & 1.1M & 32.4G \\
        \hline
        &  (64,64,64) & 4 & 2 & 2048 & 2 & 1.6 & 78.93 & 1.1M & 32.4G \\
        (C) & & &  &    & 1.4 & 1.6 & 78.22 & 1.1M & 32.4G \\
        &  &  &  &  & 3 & 2 & 78.10 & 1.1M & 32.4G \\
        &  &  &  &  & 3 & 1 & 78.96 & 1.1M & 32.4G \\
        \hline
        (D) & (64,128,256) & 4 & 2 &  2048 & 2 & 1 & 79.80 & 5.1M & 73.5G \\
        \hline
    \end{tabular*}
    \caption{Ablation of model components with PointPillars backbone on KITTI moderate Car class of \textit{val} split.}
    \label{tab:ablation_components}
\end{table}

\begin{figure}[t]
\centering
\includegraphics[width=0.95\linewidth]{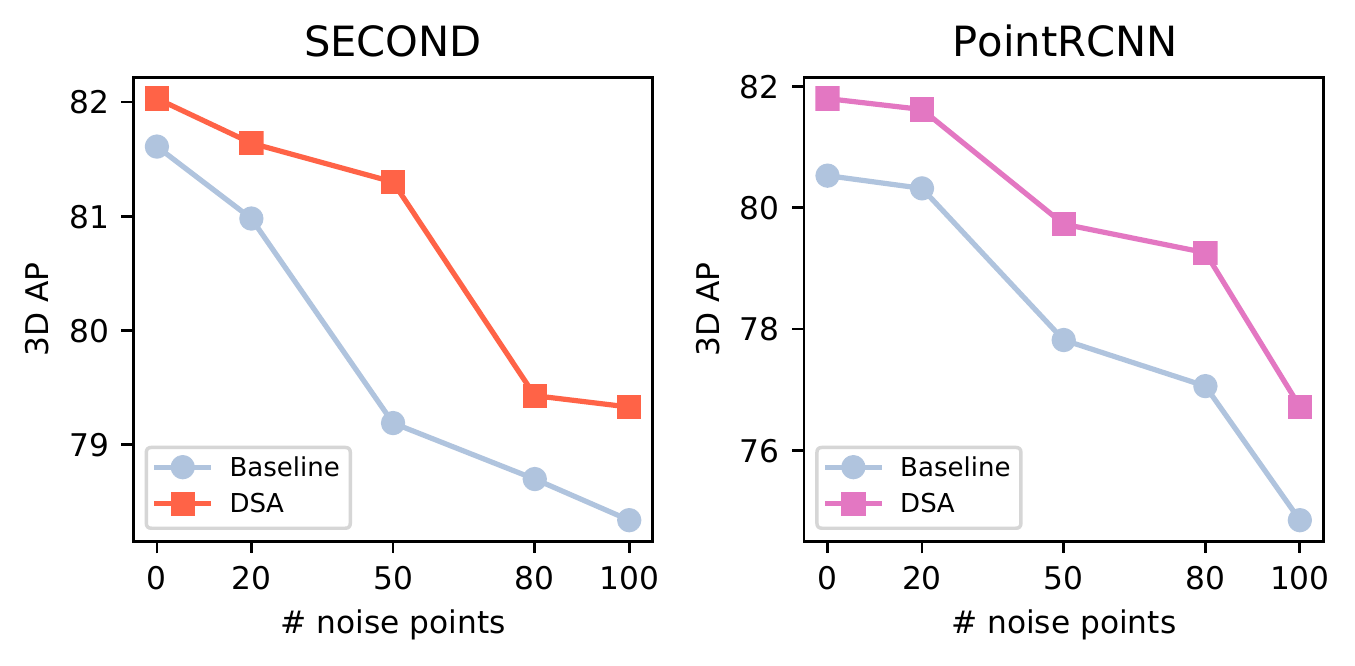}
\caption{3D AP of SECOND-DSA (orange) \& Point-RCNN-DSA (violet) vs. SECOND \& Point-RCNN baseline (light-steel-blue) for noise-points per ground-truth bounding box, varying from 0 to 100 on KITTI \textit{val} moderate}
\label{fig:graph_noise}
\end{figure}

\paragraph{Effect of noise on performance} We introduce noise points to each object similar to TANet \cite{tanet}, to probe the robustness of  representations learned. As shown in \Cref{fig:graph_noise}, self-attention-augmented models are more robust to noise than the baseline. For example, with 100 noise points added, the performance of SECOND and Point-RCNN drops by 3.3\% and 5.7\% respectively as compared to SECOND-DSA and Point-RCNN-DSA, which suffer a lower drop of 2.7\% and 5.1\% respectively.
\vspace{-0.4cm}

\begin{figure}[t]
\centering
\includegraphics[width=\linewidth]{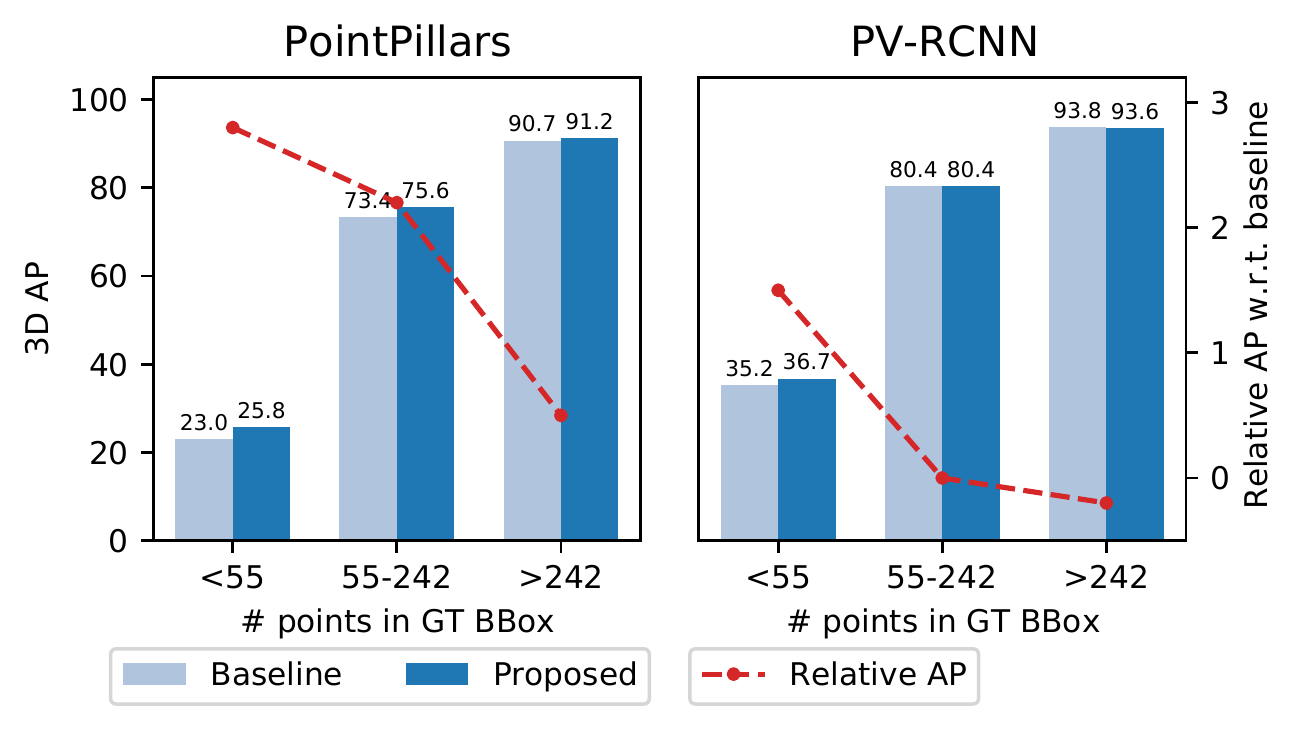}
\caption{3D AP of PointPillars-FSA, PV-RCNN-FSA and respective baselines vs.number of points in the ground-truth bounding box on KITTI \textit{val}}
\label{fig:ap_vs_numpts}
\vspace{-0.5cm}
\end{figure}

\begin{figure*}[t]
     \centering
     \includegraphics[width=\textwidth]{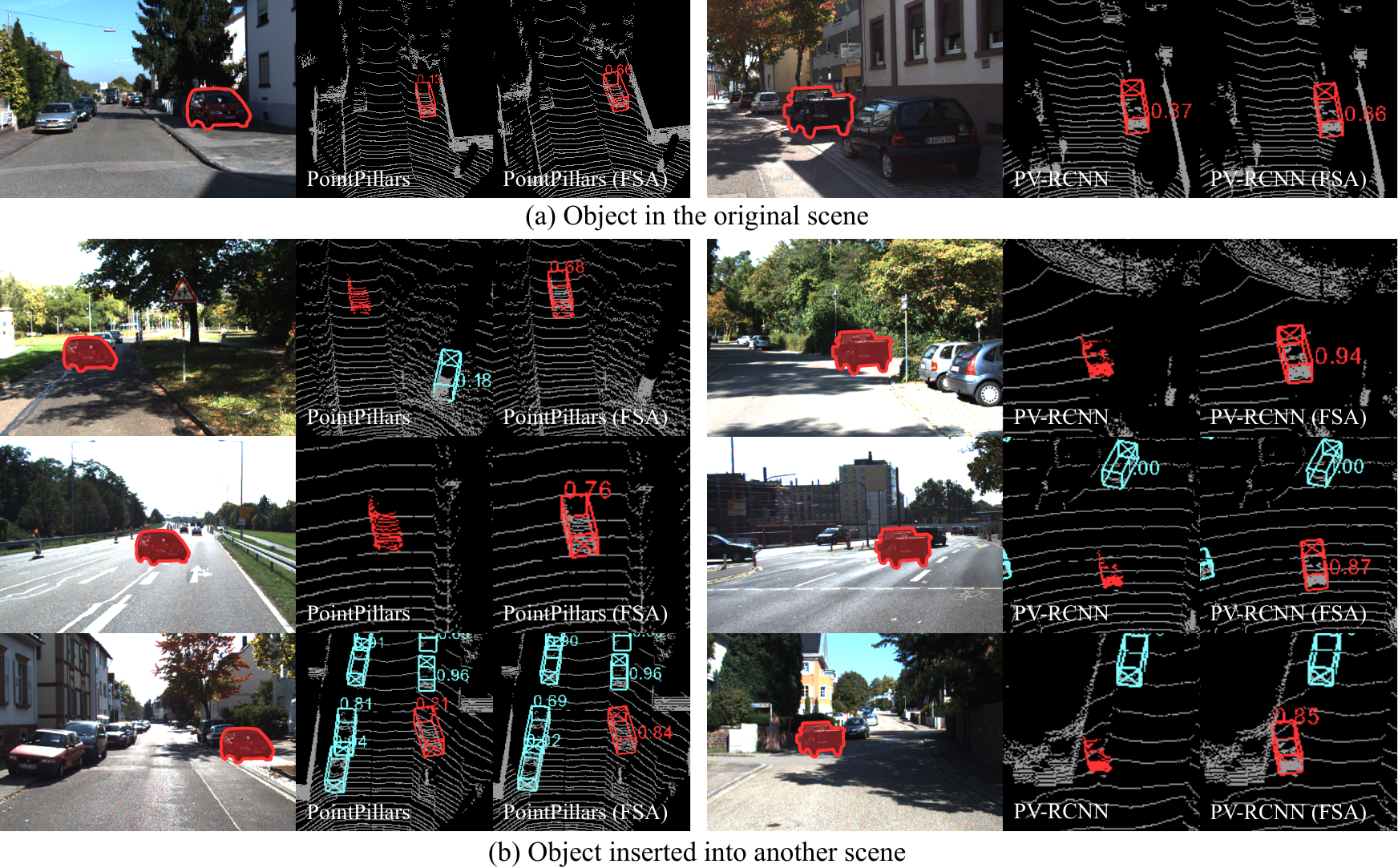}
\caption{(a) RGB images and point-clouds of cars on KITTI-\textit{val} in which addition of context via FSA had the largest increase in the detection confidence. (b) We use a simple \textit{copy and paste} method on these cars to create new point-clouds for testing our attention-based context aggregation detector for Point-Pillars and PV-RCNN backbone. We find that our FSA-based detector is more accurate and robust across scenes compared to the baseline.}
     \label{fig:aug_visualization}
\vspace{-0.5cm}
\end{figure*}

\paragraph{Effect of number of object points on performance}  We sort the cars based on the numbers of points in them in increasing order, and divide them into 3 groups based on the sorted order. Then we calculate the 3D AP across every group. As shown in \cref{fig:ap_vs_numpts}, the effect of the self-attention module becomes apparent as the number of points on the cars decreases. For objects with very few points, FSA can increase the 3D AP for PointPillars by 2.8\% and PV-RCNN by 1.5\%.

\paragraph{Qualitative results} In \Cref{fig:teaser}, we first show that our FSA-based detector identifies missed detections and eliminates false positives across challenging scenes for different backbones. 
Next, we identify objects for which addition of self-attention shows the largest increase in detection confidences as shown in \Cref{fig:aug_visualization}(a). We then copy-paste the point-clouds for these cars into different scenes. Our expectation is that the FSA is a more robust detector and can detect these examples even when randomly transplanted to different scenes. The first two rows of \Cref{fig:aug_visualization}(b) show that FSA is capable of detecting the copy-pasted car in different scenes while the baseline consistently misses them. This supports our motivation that adding contextual self-attention features to convolutional maps results in a more accurate and robust feature extractor. In the third row of \Cref{fig:aug_visualization}(b), we show cases for Point-Pillars and PV-RCNN where the orientation is flipped for our FSA-based detector even though the detection confidence remains high. We expect that this confusion occurs because FSA aggregates context from nearby high-confidence detections thereby correlating their orientations. We provide attention-visualizations and some speculation for situations in which self-attention based context aggregation is most effective in the supplementary.
\section{Conclusions}
In this paper, we propose a simple and flexible self-attention based framework to augment convolutional features with global contextual information for 3D object detection. Our proposed modules are generic, parameter and compute-efficient, and can be integrated into a range of 3D detectors. Our work explores two forms of self-attention: full (FSA) and deformable (DSA). The FSA module encodes pairwise relationships between all 3D entities, whereas the DSA operates on a representative subset to provide a scalable alternative for global context modeling. Quantitative and qualitative experiments demonstrate that our architecture systematically improves the performance of 3D object detectors.

{\small
\bibliographystyle{ieee_fullname}
\bibliography{egbib}
}

\clearpage
\begin{center}
\textbf{\large Supplementary Material for SA-Det3D: Self-Attention Based Context-Aware 3D Object Detection}
\end{center}
\renewcommand{\theequation}{\arabic{equation}}
\renewcommand{\thefigure}{\arabic{figure}}
In this document, we provide technical details and additional experimental results to supplement our main submission. We first discuss the implementation and training details used for our experiments. We then showcase the flexibility and robustness of our models through extended results on the the KITTI \cite{KITTI} dataset. We further qualitatively show the superiority of our proposed module-augmented implementations over the baseline across different challenging scenes. We also visualize the attention maps, where we observe the emergence of semantically meaningful behavior that captures the relevance of context in object detection. We finally briefly review existing standard feature extractors for 3D object detection to support our design.

\section{Network Architectures and Training Details}
\setlength{\tabcolsep}{5pt}
\begin{table}[b]
    \centering
    \begin{tabular}{c||c|c|c}
        \hline
        Backbone & Batch Size & Start LR & Max LR \\
        \hline
        PointPillars & \multirow{2}{*}{16} &
        \multirow{2}{*}{0.0003} & \multirow{2}{*}{0.003} \\
        \cline{1-1}
        SECOND & & & \\
        \hline
        Point-RCNN & \multirow{2}{*}{8} &
        \multirow{2}{*}{0.001} & \multirow{2}{*}{0.01} \\
        \cline{1-1}
        PV-RCNN & & & \\
        \hline
    \end{tabular}
    \caption{Batch size and learning rate configurations for each backbone model on KITTI benchmark}
    \label{tab:train_config}
\end{table}
In this section, we provide a more detailed description of the architectures used in our experiments. 

\subsection{Architectural details}
The detailed specification of the various layers in our FSA and DSA augmented baselines---PointPillars \cite{pointpillars}, SECOND \cite{SECOND}, Point-RCNN \cite{PointRCNN} and PV-RCNN \cite{PVRCNN}---is documented in \Cref{tab:pointpillar_archi}, \Cref{tab:second_archi}, \Cref{tab:pointrcnn_archi}, and \Cref{tab:pvrcnn}, respectively. We also provide the details of a reduced parameter baseline that aims to compare the performance of the model with similar number of parameters and FLOPs compared to their FSA and DSA counterparts.

\subsection{Experimental settings}
Additional details on encoding, training, and inference parameters are as follows. For pillar and voxel-based detection, we use absolute-position encoding for the full self-attention blocks \cite{SA}. For the \textit{test} submissions to KITTI and nuScenes official servers, we retain the full parameterization of the original baselines. For the nuScenes \textit{test} submission, we follow the configuration for PP described in \cref{tab:pointpillar_archi}, while adding the DSA module with 4 heads, 2 layers, 64-dimensional context, 2\,m deformation radius, 4096 sampled pillars and an up-sampling method described in the following subsection. For the Waymo Open dataset \textit{validation} evaluation, we use the configuration for DSA-SECOND as described in \cref{tab:second_archi}, except that we use 4096 sampled keypoints, 2\,m deformation radius and 1\,m interpolation radius. We use Pytorch \cite{Pytorch} and the recently released OpenPCDet \cite{OpenPCDet} repository for our experiments. Our models are trained from scratch in an end-to-end manner with the ADAM optimizer \cite{Adam}. The learning rates used for the different models are given in \cref{tab:train_config}. For the proposal refinement stage in two-stage networks \cite{PointRCNN}, \cite{PVRCNN}, we randomly sample $128$ proposals with 1:1 ratio for positive and negative proposals. A proposal is considered positive if it has at-least $0.55$ 3D IoU with the ground-truth boxes, otherwise it is considered to be negative. For inference, we keep the top-$500$ proposals generated from  single stage approaches \cite{pointpillars}, \cite{SECOND} and the top-$100$ proposals generated from two stage approaches \cite{PointRCNN}, \cite{PVRCNN} with a 3D IoU threshold of 0.7 for non-maximum-suppression (NMS). An NMS classification threshold of 0.1 is used to remove weak detections.

\subsection{Up-sampling for Deformable Self-Attention (DSA)}
Given the features for $m$ sampled, deformed and attended points, we explore \textit{two} up-sampling methods to distribute the accumulated structural information back to all $n$ node locations. We first test the feature propagation method proposed in PointNet++ \cite{pointnetplusplus} to obtain point features for all the original nodes. This works well for most of our experiments, especially on the KITTI and the Waymo Open Dataset. While this is simple and easy to implement, a draw-back is that the interpolation radius has to be chosen empirically. To avoid choosing an interpolation radius for the diverse classes present in the nuScenes dataset, we explore an attention-based up-sampling method as proposed in \cite{SetTransformer}. The set of $m$ points is attended to by the $n$ node features to finally produce a set of $n$ elements. This up-sampling method works well for the nuScenes dataset.
\setlength{\tabcolsep}{15pt}
\begin{table*}[t]
    \centering
    \begin{tabular*}{\textwidth}{|c||c|c|c|c|}
        \hline
        Attribute & PP \cite{pointpillars} & PP$_{red}$ & FSA-PP & DSA-PP \\
        \hline
        \hline
         \multicolumn{5}{|c|}{Layer: 2D CNN Backbone}  \\
        \hline
         Layer-nums & [3, 5, 5] & [3, 5, 5] & [3, 5, 5] & [3, 5, 5]\\
         \hline
         Layer-stride & [2, 2, 2] & [2, 2, 2] & [2, 2, 2] & [2, 2, 2] \\
         \hline
         Num-filters & [64, 128, 256] & [64, 64, 128] & [64, 64, 64] & [64, 64, 64] \\
         \hline
         Upsample-stride & [1, 2, 4] & [1, 2, 4] & [1, 2, 4] & [1, 2, 4] \\
         \hline
         Num-upsample-filters & [128, 128, 128] & [128, 128, 128] & [128, 128, 128] & [128, 128, 128] \\
         \hline
         \hline
        \multicolumn{5}{|c|}{Layer: Self-Attention}  \\
        \hline
        \hline
         Stage Added & - & - & Pillar feature & Pillar feature \\
         \hline
          Num layers & - & - & 2 & 2 \\
         \hline
          Num heads & - & - & 4 & 4 \\
         \hline
         Context Linear Dim & - & - & 64 & 64 \\
         \hline
         Num Keypoints & - & - & - & 2048 \\
         \hline
         Deform radius & - & - & - & 3.0m \\
         \hline
         Feature pool radius & - & - & - & 2.0m \\
         \hline
         Interpolation MLP Dim & - & - & - & 64 \\
         \hline
         Interpolation radius & - & - & - & 1.6m \\
         \hline
         Interpolation samples & - & - & - & 16 \\
         \hline
         
    \end{tabular*}
    \caption{Architectural details of PointPillars \cite{pointpillars}, our reduced parameter PointPillars version, proposed FSA-PointPillars and DSA-PointPillars \\ \\}
    \label{tab:pointpillar_archi}
\end{table*}
\setlength{\tabcolsep}{13.8pt}
\begin{table*}[t]
    \centering
    \begin{tabular*}{\textwidth}{|c||c|c|c|c|}
        \hline
        Attribute & SECOND \cite{SECOND} & SECOND$_{red}$ & FSA-SECOND & DSA-SECOND \\
        \hline
        \hline
        \multicolumn{5}{|c|}{Layer: 3D CNN Backbone}  \\ \cline{1-5}
        \hline
         Layer-nums in Sparse Blocks & [1, 3, 3, 3] & [1, 3, 3, 2] & [1, 3, 3, 2] & [1, 3, 3, 2]\\
         \hline
         Sparse tensor size & 128 & 64 & 64 & 64 \\
         \hline
         \hline
         \multicolumn{5}{|c|}{Layer: 2D CNN Backbone}  \\ 
        \hline
        \hline
         Layer-nums & [5, 5] & [5, 5] & [5, 5] & [5, 5]\\
         \hline
         Layer-stride & [1, 2] & [1, 2] & [1, 2] & [1, 2] \\
         \hline
         Num-filters & [128, 256] & [128, 160] & [128, 128] & [128, 128] \\
         \hline
         Upsample-stride & [1, 2] & [1, 2] & [1, 2] & [1, 2] \\
         \hline
         Num-upsample-filters & [256, 256] & [256, 256] & [256, 256] & [256, 256] \\
         \hline
         \hline
        \multicolumn{5}{|c|}{Layer: Self-Attention}  \\ 
        \hline
        \hline
        Stage Added & - & - & Sparse Tensor & Sparse Tensor \\
         \hline
          Num layers & - & - & 2 & 2 \\
         \hline
          Num heads & - & - & 4 & 4 \\
          \hline
         Context Linear Dim & - & - & 64 & 64 \\
         \hline
         Num Keypoints & - & - & - & 2048 \\
         \hline
         Deform radius & - & - & - & 4.0m \\
         \hline
         Feature pool radius & - & - & - & 4.0m \\
         \hline
         Interpolation MLP Dim & - & - & - & 64 \\
         \hline
         Interpolation radius & - & - & - & 1.6m \\
         \hline
         Interpolation samples & - & - & - & 16 \\
         \hline
         
    \end{tabular*}
    \caption{Architectural details of SECOND \cite{SECOND}, our reduced parameter SECOND version, and proposed FSA-SECOND and DSA-SECOND}
    \label{tab:second_archi}
\end{table*}
\setlength{\tabcolsep}{0.43pt}
\begin{table*}[ht]
    \centering
    \begin{tabular*}{\textwidth}{|c||c|c|c|c|}
        \hline
        \thead{Attribute} & \thead{Point-RCNN \cite{PointRCNN}} & \thead{Point-RCNN$_{red}$} & \thead{FSA-Point-RCNN} & \thead{DSA-Point-RCNN} \\
        \hline
        \hline
         \multicolumn{5}{|c|}{Layer: Multi-Scale Aggregation}  \\
        \hline
        N-Points & [4096, 1024, 256, 64] & [4096, 1024, 256, 64] & [4096, 1024, 256, 64] & [4096, 1024, 128, 64] \\
        \hline
         Radius &  \makecell{[0.1, 0.5], [0.5, 1.0], \\ {[1.0, 2.0], [2.0, 4.0]}} & \makecell{[0.1, 0.5], [0.5, 1.0], \\ {[1.0, 2.0], [2.0, 4.0]}} & \makecell{[0.1, 0.5], [0.5, 1.0], \\ {[1.0, 2.0], [2.0, 4.0]}} & \makecell{[0.1, 0.5], [0.5, 1.0], \\ {[1.0, 2.0], [2.0, 4.0]}} \\
         \hline
         N-samples & [16, 32] & [16, 32] & [16, 32] & [16, 32] \\
         \hline
         MLPs & \makecell{[16, 16, 32], [32, 32, 64], \\ {[64, 64, 128], [64, 96, 128]} \\ {[128, 196, 256], [128, 196, 256]} \\ {[256, 256, 512], [256, 384, 512]}} & \makecell{[16, 32], [32, 64], \\ {[64, 128], [64, 128]} \\ {[128, 256], [128, 256]} \\ {[256, 512], [256, 512]}} & \makecell{[16, 32], [32, 64], \\ {[64, 128], [64, 128]} \\ {[128, 256], [128, 256]} \\ {[256, 512], [256, 512]}} & \makecell{[16, 32], [32, 64], \\ {[64, 128], [64, 128]} \\ {[128, 256], [128, 256]} \\ {[256, 512], [256, 512]}} \\
         \hline
         FP-MLPs  &  \makecell{[128, 128], \\ {[256, 256]}, \\ {[512, 512]}, \\ {[512, 512]}} & \makecell{[128, 128], \\ {[128, 128]}, \\ {[128, 128]}, \\ {[128, 512]}} & \makecell{[128, 128], \\ {[128, 128]}, \\ {[128, 128]}, \\ {[128, 128]}} & \makecell{[128, 128], \\ {[128, 128]}, \\ {[128, 128]}, \\ {[128, 128]}}    \\
         \hline
         \hline
        \multicolumn{5}{|c|}{Layer: Self-Attention}  \\
        \hline
        \hline
        Stage Added & - & - & MSG-3 and MSG-4 & MSG-3 and MSG-4 \\
         \hline
          Num layers & - & - & 2 & 2 \\
         \hline
          Num heads & - & - & 4 & 4 \\
         \hline
         Context Linear Dim & - & - & 64 & 64 \\
         \hline
         Num Keypoints & - & - & - & (128, 64) \\
         \hline
         Deform radius & - & - & - & (2.0, 4.0)m \\
         \hline
         Feature pool radius & - & - & - & (1.0, 2.0)m \\
         \hline
         Interpolation MLP Dim & - & - & - & (64, 64) \\
         \hline
         Interpolation radius & - & - & - & (1.0, 2.0)m \\
         \hline
         Interpolation samples & - & - & - & (16, 16) \\
         \hline 
    \end{tabular*}
    \caption{Architectural details of Point-RCNN \cite{PointRCNN}, our reduced parameter Point-RCNN version, proposed FSA-Point-RCNN and DSA-Point-RCNN \\ \\  \\  \\ \\ \\ \\ \\  \\}
    \label{tab:pointrcnn_archi}
    \vspace{-3.5cm}
\end{table*}
\setlength{\tabcolsep}{15.2pt}
\begin{table*}[ht]
    \centering
    \begin{tabular*}{\textwidth}{|c||c|c|c|}
        \hline
        Attribute & PV-RCNN \cite{PVRCNN} & FSA-PVRCNN & DSA-PVRCNN \\
        \hline
        \hline
        \multicolumn{4}{|c|}{Layer: 3D CNN Backbone}  \\ \cline{1-4}
        \hline
         Layer-nums in Sparse Blocks & [1, 3, 3, 3]  & [1, 3, 3, 2] & [1, 3, 3, 3]\\
         \hline
         Sparse tensor size & 128 & 64 & 128 \\
         \hline
         \hline
         \multicolumn{4}{|c|}{Layer: 2D CNN Backbone}  \\ 
        \hline
        \hline
         Layer-nums & [5, 5] & [5, 5] & [5, 5]\\
         \hline
         Layer-stride & [1, 2] & [1, 2] & [1, 2] \\
         \hline
         Num-filters & [128, 256] & [128, 128] & [128, 256] \\
         \hline
         Upsample-stride & [1, 2] & [1, 2] & [1, 2] \\
         \hline
         Num-upsample-filters &  [256, 256] & [256, 256] & [256, 256] \\
         \hline
         \hline
        \multicolumn{4}{|c|}{Layer: Self-Attention}  \\ 
        \hline
        \hline
        Stage Added & - & Sparse Tensor and VSA & VSA \\
         \hline
          Num layers  & - & 2 & 2 \\
         \hline
          Num heads & - & 4 & 4 \\
          \hline
         Context Linear Dim & - & 128 & 128 \\
         \hline
         Num Keypoints & -  & - & 2048 \\
         \hline
         Deform radius &  - & - & \makecell{[0.4, 0.8], [0.8, 1.2], \\ {[1.2, 2.4], [2.4, 4.8]}} \\
         \hline
         Feature pool radius  & - & - & Multi-scale: (0.8, 1.6)m \\
         \hline
         Interpolation MLP Dim  & - & - & Multi-scale: (64, 64) \\
         \hline
         Interpolation radius  & - & - & Multi-scale: (0.8, 1.6)m \\
         \hline
         Interpolation samples  & - & - & Multi-scale: (16, 16) \\
         \hline
         
    \end{tabular*}
    \caption{Architectural details of PV-RCNN \cite{PVRCNN}, and proposed FSA-PVRCNN and DSA-PVRCNN}
    \label{tab:pvrcnn}
\end{table*}
\section{Detailed Results}
We provide additional experimental details on the validation split of the KITTI \cite{KITTI} data-set in this section. In \Cref{tab:kitti_val_cyclist}, we first show the 3D and BEV AP for moderate difficulty on the \textit{Cyclist} class for PV-RCNN and its variants. The table shows that both our proposed modules improve on the baseline results. This showcases the robustness of our approach in also naturally benefiting smaller and more complicated objects like cyclists. We then proceed to list the 3D AP and BEV performances with respect to distance from the ego-vehicle in \Cref{table:distance}. We find that the proposed blocks especially improve upon detection at further distances, where points become sparse and context becomes increasingly important. These results hold especially for the \textit{cyclist} class---as opposed to the \textit{car} class, which shows that context is possibly more important for smaller objects with reduced number of points available for detection. In \Cref{tab:kitti_val_details}, we provide results for all three difficulty categories for the \textit{car} class. We see consistent improvements across backbones with various input modalities on the \textit{hard} category. This is consistent with our premise that samples in the hard category can benefit more context information of surrounding instances. We also note that PointPillars \cite{pointpillars}, which loses a lot of information due to pillar-based discretization of points, can supplement this loss with fine-grained context information.
\section{Qualitative Results} 
\subsection{Comparison with Baseline}
\begin{figure*}[ht]
    \centering
    \includegraphics[width=\textwidth]{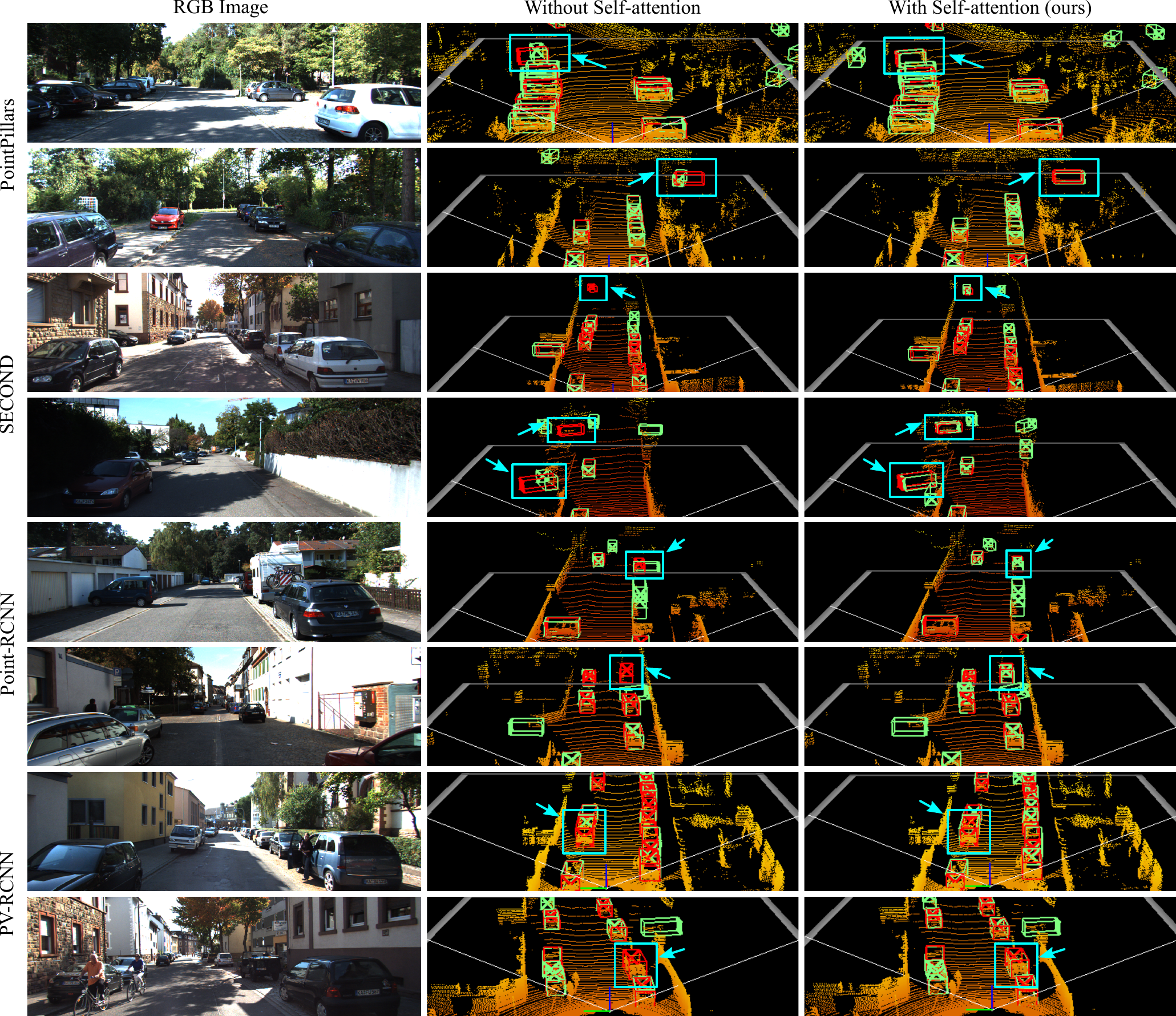}
    \caption{Qualitative comparisons of our proposed approach with the baseline on the KITTI validation set. \textit{Red} represents Ground-Truth bounding box while \textit{Green} represents detector outputs. From left to right: RGB images of scenes; Baseline performance across state-of-the-art detectors PointPillars \cite{pointpillars}, SECOND \cite{SECOND}, Point-RCNN \cite{PointRCNN} and PV-RCNN \cite{PVRCNN}; Performance of proposed FSA module-augmented detectors. Viewed best when enlarged.}
    \label{fig:qual_compare}
\end{figure*}

In this section, we provide additional qualitative results across challenging scenarios from real-world driving scenes and compare them with the baseline performance (see \Cref{fig:qual_compare}). The ground-truth bounding boxes are shown in \textit{red}, whereas the detector outputs are shown in \textit{green}. We show consistent improvement in identifying missed detections across scenes and with different backbones including PointPillars \cite{pointpillars}, SECOND \cite{SECOND}, Point-RCNN \cite{PointRCNN} and PV-RCNN \cite{PVRCNN}. We note that we can better refine proposal bounding box orientations with our context-aggregating FSA module (Rows 1, 2, and 4). We also note that cars at distant locations can be detected by our approach (Rows 3, 4 and 6). Finally we analyze that cars with slightly irregular shapes even at nearer distances are missed by the baseline but picked up by our approach (Rows 7 and 8). 

\subsection{Visualization of Attention Weights}
\begin{figure*}[h]
    \centering
    \includegraphics[width=\textwidth]{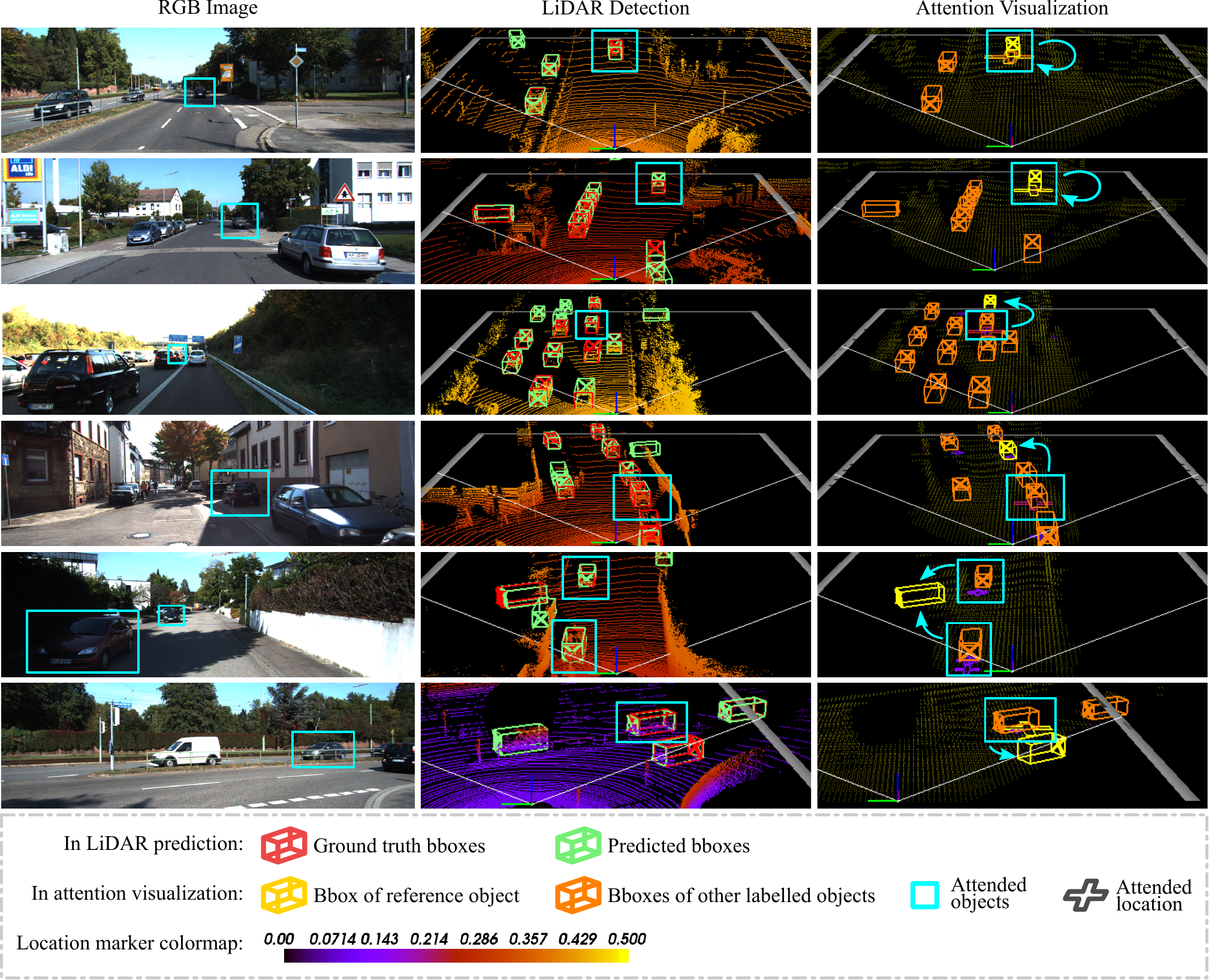}
    \caption{Visualization of attention maps produced by our proposed FSA-variant on SECOND \cite{SECOND} backbone. We analyze the implications of the produced attention maps in Section 3.2.}
    \label{fig:vis_attn}
\end{figure*}

We also visualize the attention weights for FSA-variant for the SECOND \cite{SECOND} backbone in \Cref{fig:vis_attn}. In this implementation, voxel features down-sampled by 8-times from the point-cloud space are used to aggregate context information through pairwise self-attention. We first visualize the voxel space, where the center point of each voxel is represented as a yellow point against the black scene-background. We next choose the center of a ground-truth bounding box as a reference point. We refer this bounding box as the reference bounding box. The reference bounding box is shown in \textit{yellow}, and the rest of the labeled objects in the scene are shown in \textit{orange}. We next visualize the attention weights across all the voxel centers with respect to the chosen reference bounding box center. Of the 4 attention maps produced by the 4 FSA-heads, we display the attention map with the largest activation in our figures. We find that attention weights become concentrated in small areas of the voxel-space. These voxel centers are called attended locations and are represented by a thick cross in our visualizations. The color of the cross represents the attention weight at that location and the scale of attention weights is represented using a colorbar. The size of the cross is manipulated manually by a constant factor. In an effort to improve image-readability, we connect the chosen reference object to the other labelled objects in the scene that it pays attention to (with blue boxes and blue arrows) as inferred from the corresponding attended locations while aggregating context information. 
\setlength{\tabcolsep}{16.7pt}
\begin{table*}[h]
\centering
    \begin{tabular*}{\textwidth}{|c|c|cc|ccc|}
        \hline
        Model & Modality & Params & GFLOPs & \multicolumn{3}{c|}{Car 3D AP} \\ 
        & & (M) & & Easy & Moderate & Hard \\
        \hline
        PP \cite{pointpillars} & BEV & 4.8 & 63.4 & 87.75 & 78.39 & 75.18
         \\
         PP$_{red}$ & BEV & 1.5 & \textbf{31.5} & 88.09 & 78.07 & 75.14
        \\
         PP-DSA & BEV & 1.1 & 32.4 & 89.37 & 78.94 & 75.99
        \\
        PP-FSA & BEV & \textbf{1.0} & 31.7 & \textbf{90.10} & \textbf{79.04} & \textbf{76.02}
        \\
        \hline
    SECOND \cite{SECOND} & Voxel & 4.6 & 76.7 & 90.55 & 81.61 & 78.61
         \\
         SECOND$_{red}$ & Voxel & 2.5 & \textbf{51.2} & 89.93 & 81.11 & 78.30
        \\
         SECOND-DSA & Voxel & 2.2 & 52.6 & \textbf{90.70} & \textbf{82.03} & \textbf{79.07}
        \\
        SECOND-FSA & Voxel & \textbf{2.2} & 51.9 & 89.05 & 81.86 & 78.84
        \\
        \hline
        Point-RCNN \cite{PointRCNN} & Points & 4.0 & 27.4 & \textbf{91.94} & 80.52 & 78.31
         \\
         Point-RCNN$_{red}$ & Points & 2.2 & 24.1 & 91.47 & 80.40 & 78.07
        \\
         Point-RCNN-DSA & Points & \textbf{2.3} & \textbf{19.3} & 91.55 & 81.80 & 79.74
        \\
        Point-RCNN-FSA & Points & 2.5 & 19.8 & 91.63 & \textbf{82.10} & \textbf{80.05}
        \\
        \hline
    \end{tabular*}
    \caption{Detailed comparison of 3D AP with baseline on KITTI \textit{val} split with 40 recall positions}
    \label{tab:kitti_val_details}
\end{table*}

\setlength{\tabcolsep}{1pt}
\begin{table}[h]
    \centering
    \begin{tabular}{c||cc}
        \hline
        & {3D} & {BEV}  \\
        \hline
        PV-RCNN \cite{PVRCNN} & 70.38 & 74.5 \\
        PV-RCNN + DSA & \bf{73.03} &\bf{75.45}  \\
        PV-RCNN + FSA & 71.46 & 74.73  \\
        \hline
    \end{tabular}
    \caption{Performance comparison for moderate difficulty cyclist class on KITTI \textit{val} split.}
    \label{tab:kitti_val_cyclist}
\end{table}
\setlength{\tabcolsep}{1.4pt}
\setlength{\tabcolsep}{4pt}
\begin{table}[h]
\begin{center}
\begin{tabular}{c||c||cccc}
\hline
Distance & Model & Car & Cyclist & Pedestrian \\
\hline
 \multirow{3}{*}{0-30m} & PV-RCNN \cite{PVRCNN} & 91.71 & 73.76 & 56.82 \\ 
  & DSA & 91.65 & \bf{74.89} & 59.61  \\
  & FSA & \bf{93.44} & 74.10 & \textbf{61.65} \\
 \hline
 \multirow{3}{*}{30-50m} & PV-RCNN \cite{PVRCNN} & 50.00 & 35.15 & - \\ 
 & DSA & 52.02 & \bf{47.00} & -  \\
 & FSA & \bf{52.76} & 39.74 & - \\
\hline
\end{tabular}
\end{center}
\caption{Comparison of nearby and distant-object detection on the moderate level of KITTI \textit{val} split with AP calculated by 40 recall positions}
\label{table:distance}
\end{table}
\setlength{\tabcolsep}{1.4pt}

In our paper, we speculate that sometimes for true-positive cases, CNNs (which are essentially a pattern matching mechanism) detect a part of the object but are not very confident about it. This confidence can be increased by looking at nearby voxels and inferring that the context-aggregated features resemble a composition of parts. We therefore first ask the question if our FSA module can adaptively focus on its own local neighbourhood. We show in Rows 1 and 2 of \Cref{fig:vis_attn} that it can aggregate local context adaptively. We also hypothesize that, for distant cars, information from cars in similar lanes can help refine orientation. We therefore proceed to show instances where a reference bounding box can focus on cars in similar lanes, in Rows 3 and 4 of \Cref{fig:vis_attn}. We also show cases where FSA can adaptively focus on objects that are relevant to build structural information about the scene in Rows 5 and 6 of \Cref{fig:vis_attn}. Our visualizations thus indicate that semantically meaningful patterns emerge through the self-attention based context-aggregation module.

\section{Standard Feature Extractors for 3D Object Detection}
In this section, we briefly review the standard feature extractors for 3D object detection to motivate our design. 2D and 3D convolutions have achieved great success in processing pillars \cite{pointpillars} and voxel grids \cite{SECOND} for 3D object detection. Point-wise feature learning methods like PointNet++ \cite{pointnetplusplus} have also been successful in directly utilizing sparse, irregular points for 3D object detection \cite{PointRCNN}. 

Given a set of vectors $\{x_1, x_2,...x_n\}$, which can represent pillars, voxels or points, with $x_i \in R^C$, one can define a function $f:\mathcal{X} \xrightarrow[]{} R^{C'}$ that maps them to another vector. In this case, standard convolution at location $\hat{p}$ can be formulated as:
\begin{equation}
f(\hat{p}) = \sum_{l\in\Omega_1}x_{\hat{p}+l}w_l
\end{equation}
where $w$ is a series of $C'$ dimensional weight vectors with kernel size $2m+1$ and $\Omega_{1}=[l\in(-m,...m)]$ representing the set of positions relative to the kernel center. Similarly, a point-feature approximator at $\hat{p}$ can be formulated as:
\begin{equation}
f(\hat{p}) = \max_{l\in\Omega_2}h(x_l)
\end{equation}
where $h$ is a $C'$ dimensional fully connected layer, $\max$ denotes the max-pooling operator and $\Omega_2$ denotes the $k$-nearest neighbors of $\hat{p}$. 
The operator $f$ thus aggregates features with pre-trained weights, $h$ and $w$, from nearby locations.

\paragraph{Limitations}  One of the disadvantages of this operator is that weights are fixed and cannot adapt to the content of the features or selectively focus on the salient parts. Moreover, since the number of parameters scales linearly with the size of the neighborhood to be processed, long range feature-dependencies can only be modeled by adding more layers, posing optimization challenges for the network. Since useful information for fine-grained object recognition and localization appears at both global and local levels of a point-cloud, our work looks for more effective feature aggregation mechanisms.

\end{document}